\documentclass[pdflatex,sn-apa]{sn-jnl}

\jyear{2023}%

\theoremstyle{thmstyleone}%
\usepackage{graphicx}
\usepackage{caption}
\usepackage{adjustbox}
\usepackage{subcaption}
\usepackage{multirow}

\newcommand{\bfw}{\mathbf{w}}

\newcommand{\bfa}{\mathbf{a}}

\newcommand{\bff}{\mathbf{f}}
\newcommand{\bfy}{\mathbf{y}}

\newcommand{\bfA}{\mathbf{A}}
\newcommand{\bfD}{\mathbf{D}}
\newcommand{\bfW}{\mathbf{W}}
\newcommand{\bfS}{\mathbf{S}}

\newcommand{\bfP}{\mathbf{P}}

\newcommand{\bftheta}{\boldsymbol{\theta}}

\theoremstyle{thmstyletwo}%

\theoremstyle{thmstylethree}%

\begin{document}

\title[ENC: Improving the Training of GNNs on Node Classification]{Every Node Counts: Improving the Training of Graph Neural Networks on Node Classification}


\author*[1]{\fnm{Moshe} \sur{Eliasof}}\email{eliasof@post.bgu.ac.il}

\author[2]{\fnm{Eldad} \sur{Haber}}\email{eldadHaber@gmail.com}

\author[1]{\fnm{Eran} \sur{Treister}}\email{erant@cs.bgu.ac.il}

\affil*[1]{\orgdiv{Computer Science Department}, \orgname{Ben-Gurion University of the Negev},\orgaddress{ \city{Beer-Sheva}, \country{Israel}}}

\affil[2]{\orgdiv{Department of Earth, Ocean and Atmospheric Sciences}, \orgname{University of British Columbia}, \orgaddress{\city{Vancouver}, \country{Canada}}}


\abstract{Graph Neural Networks (GNNs) are prominent in handling sparse and unstructured data efficiently and effectively. Specifically, GNNs were shown to be highly effective for node classification tasks, where labelled information is available for only a fraction of the nodes. Typically, the optimization process, through the objective function, considers only labelled nodes while ignoring the rest. In this paper, we propose novel objective terms for the training of GNNs for node classification, aiming to exploit all the available data and improve accuracy. Our first term seeks to maximize the mutual information between node and label features, considering both labelled and unlabelled nodes in the optimization process. Our second term promotes anisotropic smoothness in the prediction maps.
Lastly, we propose a cross-validating gradients approach to enhance the learning from labelled data.
Our proposed objectives are general and can be applied to various GNNs and require no architectural modifications. Extensive experiments demonstrate our approach using popular GNNs like GCN, GAT and GCNII, reading a consistent and significant accuracy improvement on 10 real-world node classification datasets.}



\maketitle

\section{Introduction}
\label{sec:intro}
The field of Graph Neural Networks (GNNs) has gained large popularity in recent years \citep{scarselli2008graph, bruna2013spectral,defferrard2016convolutional,kipf2016semi,bronstein2017geometric} in a variety of fields and applications such as computer graphics and vision \citep{acnn_boscaini,monti2017geometric,wang2018dynamic,hanocka2019meshcnn,eliasof2020diffgcn}, Bioinformatics \citep{Strokach2020,jumper2021highly}, node classification \citep{kipf2016semi,velickovic2018graph,chen20simple} and others.
In the context of node classification, many methods \citep{kipf2016semi,velickovic2018graph,chen20simple,zhou2021dirichlet} and others, consider the semi-supervised setting, where only a small part of the nodes are labelled, e.g., 20 labelled nodes per class in the Cora \cite{sen2008collective} dataset, translating to 5\% labelled nodes. Others consider the fully-supervised setting, where typically  \citep{Pei2020Geom-GCN:,musae} 48\% of the nodes are labelled.
The recipes of the various methods share a common factor, which is the training procedure. While the aforementioned methods propose novel architectures, aggregation schemes or network dynamics, they yet require the minimization of the cross-entropy loss of \emph{labelled} node predictions. This gives rise to the research question --  it is beneficial to also consider the unlabelled information and specifically the predictions of unlabelled nodes in the training procedure, and how?

\begin{figure}
    \centering
    \begin{subfigure}{0.23\textwidth}
    \includegraphics[width=1\textwidth]
    {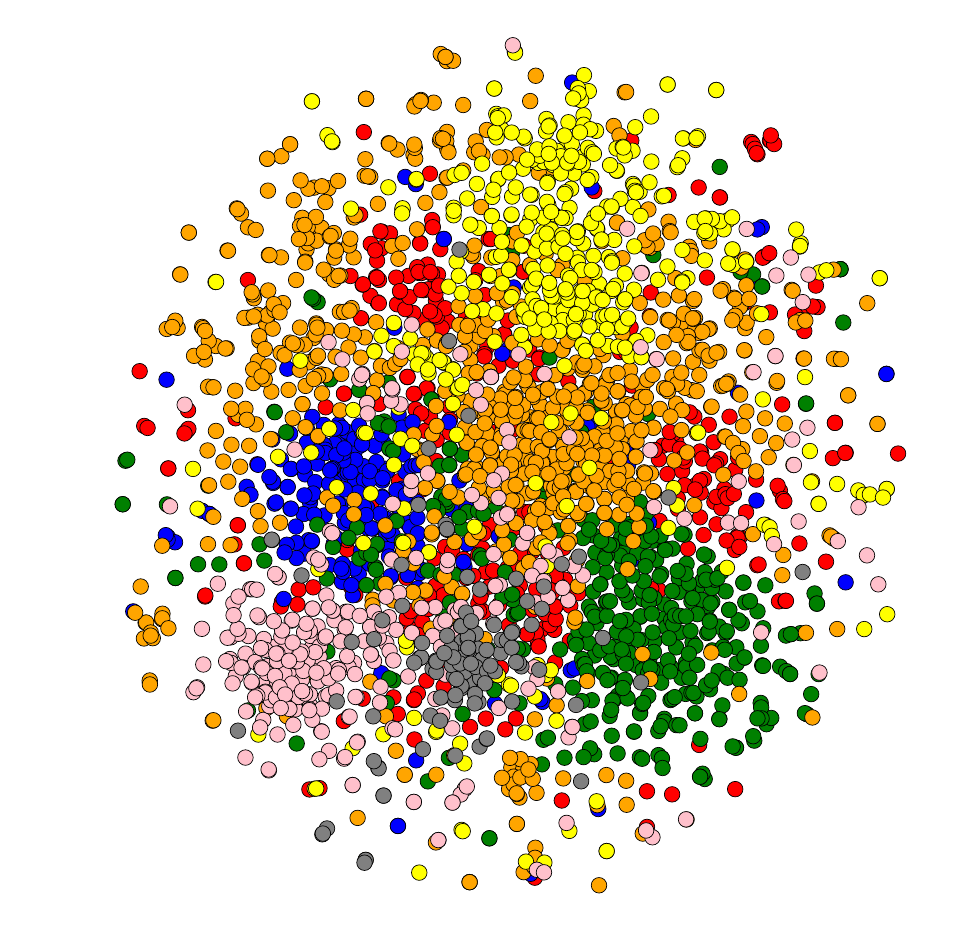}
    \caption{}
    \end{subfigure}
    \begin{subfigure}{0.3\textwidth}
    \includegraphics[width=1\textwidth]{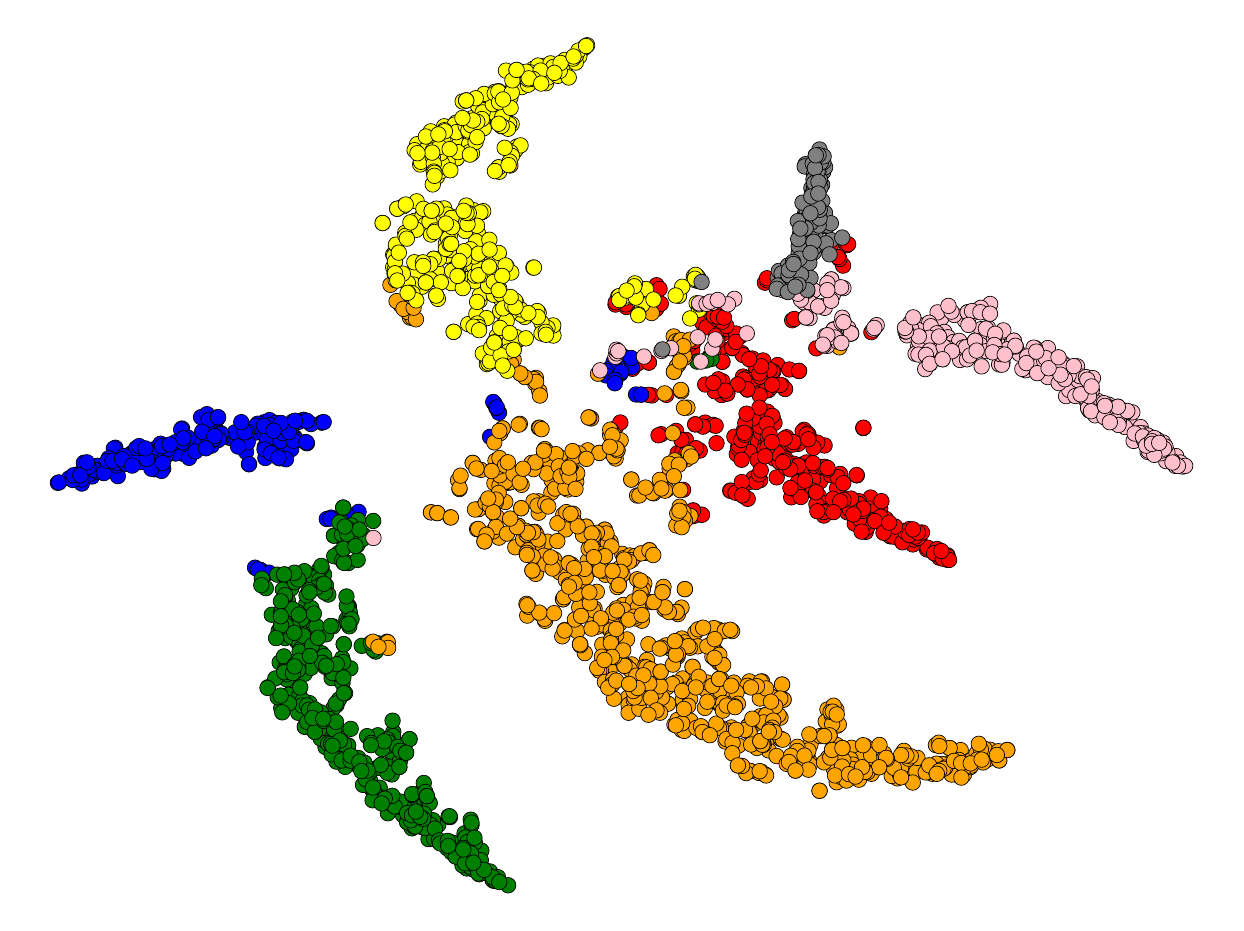}
        \caption{}
    \end{subfigure}
    \begin{subfigure}{0.3\textwidth}
    \includegraphics[width=1\textwidth]{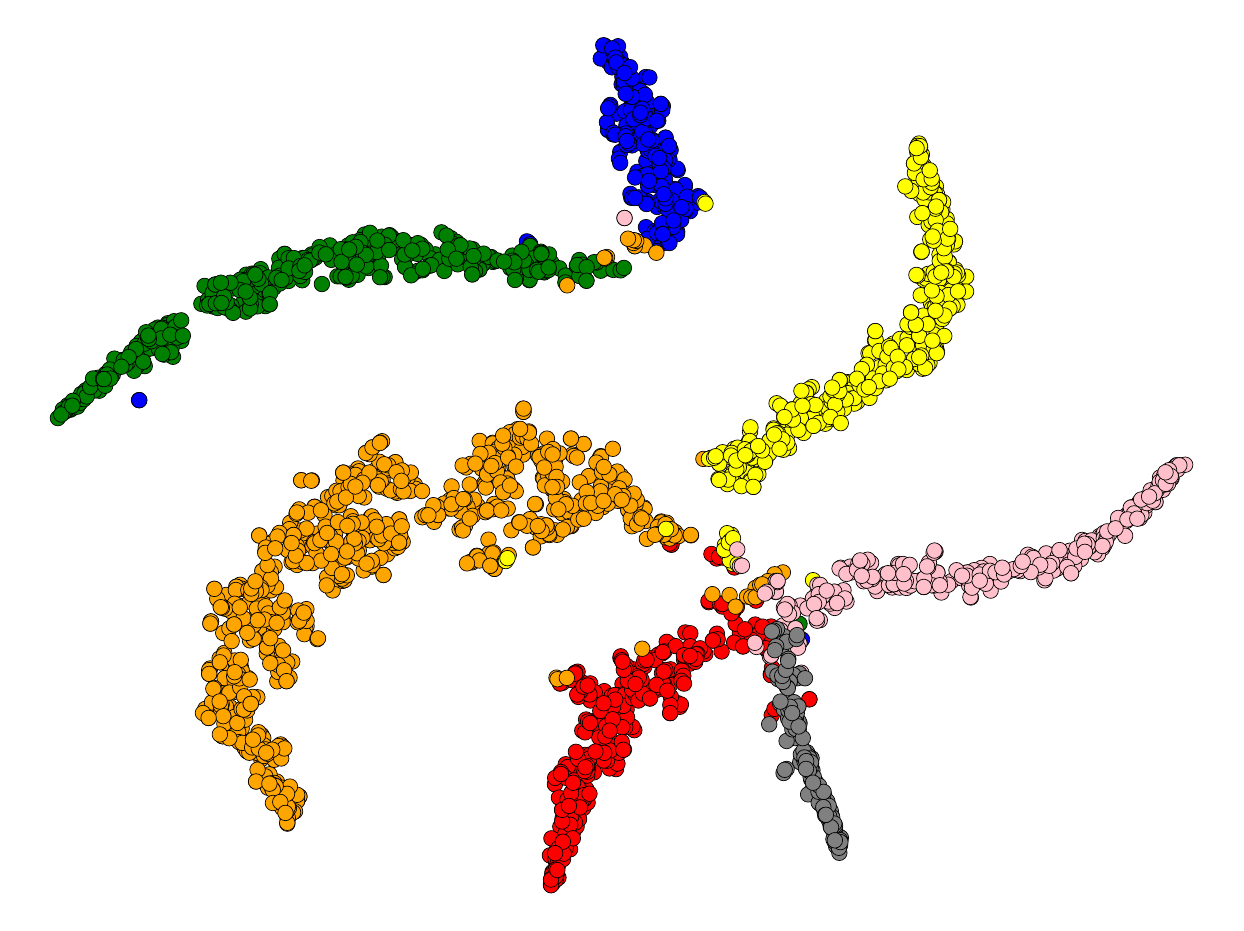}
        \caption{}
    \end{subfigure}
    \caption{t-SNE embeddings of (a) Cora input features, (b) GCN predictions, (c) ENC-GCN predictions. Our ENC-GCN improves the node classification accuracy by 4.3\% compared to GCN. Zoom in for a better view.}
    \label{fig:tsne}
\end{figure}

This question is often treated by requiring the consistency of the predicted node classification with respect to adversarial perturbations of the data \citep{feng2019graph_adversarialtrain,suresh2021adversarial}, Laplacian based regularization \citep{yang2021rethinking} of the node predictions, label entropy minimization \citep{sun2019infograph,bo2022regularizing_aaai22} and label propagation methods \citep{zhou2003learning,wang2020unifying,dong2021equivalence}. While the considered methods show significant improvement over the baseline, we identify that there are limitations that can be further relieved. For instance, demanding smoothness according to the Laplacian may be useful in the case of homophilic (following the definition from \cite{Pei2020Geom-GCN:}) datasets like Cora, Citeseer and Pubmed. However, for datasets like Cornell, Texas and Wisconsin, that have a low homophily score (i.e., heterophilic datasets), such a demand can result in degraded accuracy, as we show in Section \ref{sec:ablation}. An adversarial perturbation of the data is highly dependant on the perturbation policy, and requires additional computational costs due to the comparison of two or more forward computations of the network. Simple label entropy minimization can gravitate the network towards predicting a dominant class in the labelled data, which may not reflect the real distribution of the data. 

In this paper we propose to incorporate both labelled and unlabelled nodes in the objective functions through the perspectives of Mutual Information (MI) maximization, Total Variation (TV) regularization, as well as a Cross-Validating Gradients (CVG) approach that improves training from labelled nodes. We find that augmenting the standard cross-entropy objective function with unlabelled nodes information and enhanced labelled node information leads to a consistent improvement across all considered datasets and experiments, showing that the information from \textit{every node counts} to improve accuracy. We therefore call our method ENC. An example of the obtained node classification of our ENC approach compared to a baseline GCN is given in \ref{fig:tsne}.
Our contributions are as follows:
\begin{itemize}
    \item We propose a Mutual Information based objective function that aims to maximize prediction information between all nodes and classes.
    \item A Total Variation guided objective is proposed 
    to promote node class predictions that adhere to the natural boundaries of the graph signal.
    \item A Cross-Validating Gradients approach is introduced as a stochastic measure to improve the gradient direction from labelled nodes.
    \item Through a series of extensive experiments we demonstrate the added value of our method, reading a consistent improvement on all datasets and baselines, that is in line with current state-of-the-art methods.
\end{itemize}

\section{Related Work}
\label{sec:related}
\subsection{Graph Neural Networks}
Traditionally, GNNs are categorized into spectral \citep{bruna2013spectral} 
and spatial \citep{defferrard2016convolutional, kipf2016semi,simonovsky2017dynamic,gilmer2017neural,monti2017geometric} 
methods. Most of those can be implemented using the Message-Passing Neural Network mechanism by \cite{gilmer2017neural}, where each node aggregates features (messages) from its neighbours, according to some policy. For instance, GCN \citep{kipf2016semi} and ChebNet \citep{defferrard2016convolutional} polynomials of the graph Laplacian to parameterize the convolution operator. Works like GraphSAGE \citep{hamilton2017inductive}, PAN \citep{ma2020path} and pathGCN \citep{eliasof2022pathgcn} learn multi-hop convolutional kernels. Attention based methods like GAT \citep{velickovic2018graph}, SuperGAT \citep{kim2020findSuperGAT} and GATv2 \citep{brody2021attentive} learn a non-negative score of the graph edges to perform local propagation of node features. Computer vision oriented methods like DGCNN  \citep{wang2018dynamic} constructs a k-nearest-neighbours graph from point-clouds and dynamically updates it, and MoNet \citep{monti2017geometric} learns a Gaussian mixture model to weight the edges of meshes for shape analysis tasks. A shared quality of the aforementioned methods is the training scheme. While all methods focus on minimizing a problem designated loss (e.g., the cross-entropy loss function for node or graph classification) with respect some labelled data, there is no consideration of the unlabelled data. In this paper we focus on the incorporation of unlabelled data to improve the training of GNNs on the node-classification task.

\subsection{Improved training of Graph Neural Networks}
The study of improved training of neural networks, and in particular of GNNs is concerned with creating different training policies and losses. Perhaps the most basic and common remedy for training on the typically small datasets like Cora, Citeseer and Pubmed is the incorporation of Dropout \citep{srivastava2014dropout} after every GNN layer, which has become a standard practice \citep{kipf2016semi,xu2018how,chen20simple,zhou2021dirichlet}. Other important tools that improve training are realized by randomly alternating the data rather than the neural units of the GNN. For instance, \cite{Rong2020DropEdge:} suggest
DropEdge, a method that randomly drops graph edges, and \cite{do2021graph_dropnode} propose DropNode -- a method that randomly removes graph nodes. 
Other methods like PairNorm \citep{Zhao2020PairNorm:} propose adding node features normalization, which also helps to alleviate the over-smoothing phenomenon in GNNs discussed in \cite{measuringoversmoothing, cai2020note}.
Another approach is the Mixup \citep{zhang2017mixup,verma2019manifold} technique that enriches the learning data, and has shown success in image classification tasks. Following that, works like GraphMix \citep{verma2021graphmix} proposed an interpolation-based regularization method by parameter sharing of GNNs and point-wise convolution, and G-Mixup \citep{gmixupicml2022} proposes an augmentation technique based on graph generators for graph classification tasks.
The methods above were shown to improve the training in GNNs. However, they do not consider the information of the unlabelled nodes during the training. 

Recent methods that consider both labelled and unlabelled data in the context of improved GNN training include InfoGraph \citep{sun2019infograph} that learns a discriminative network for graph classification tasks. For graph classification, \cite{you2020graph} suggest to utilize unlabelled data through contrastive learning and data augmentations. Moreover, \cite{bo2022regularizing_aaai22} propose consistency-diversity augmentations for node and graph classification tasks, and \cite{wang2021mixup} suggest a mixup based method that considers all available data. Our work differs from the above as follows. First, we utilize an information maximization approach, with a class entropy balancing term in Eq. \eqref{eq:mi_loss}. Second, we propose a total-variation smoothing loss that is adherent to the natural edges of the graph signal, and is shown in Section \ref{sec:ablation} and Table \ref{table:TVvsLaplacianReg} to obtain improved accuracy compared to a Laplacian smoothing regularization as in \cite{yang2021rethinking} and a standard total-variation loss that was utilized in GNNs \citep{jin2020graph,liu2021elastic,liao2021information,wang2021confident}.
Third, we propose a cross-validating gradients approach to further improve learning from labelled nodes on top of the standard cross-entropy loss.

\subsection{Mutual Information in Neural Networks}
The concept of Mutual Information in machine learning tasks was originally used for image and volume alignment by \cite{wells1996multi,viola1997alignment}. Recently, it was implemented into CNNs by the seminal Deep InfoMax \citep{hjelm2018learning} and into GNNs by \cite{deepgraphInfomax}, where unsupervised learning tasks are considered, by defining some task that is defined by the data (e.g., signal reconstruction) and enforcing information maximization between inputs and their reconstruction or predictions.
This concept was found to be useful in a wide array of applications, 
from unsupervised image semantic segmentation \citep{ouali2020autoregressive,mirsadeghi2021unsupervised} to unsupervised graph related tasks \citep{deepgraphInfomax,peng2020graph}. 
In this paper we show that this concept  significantly improves the overall performance of GNNs when labelled information is partially available, without any architectural changes.
In this paper we choose the 'standard' mutual information, reference to f-mutual information about talk about other possibilities for future work.

\section{Notations and Technical Background}
\label{sec:notations}
\paragraph{Notations.} We now provide the notations that will be used throughout this paper.
Let us denote an undirected graph defined by the tuple $\cal G=({\cal V},{\cal E})$  where $\cal V$ is a set of $n$ nodes and $\cal E$ is a set of $m$ edges. We define the neighbourhood of the $i$-th node by $\mathcal{N}_i = \{j \mid (i,j) \in \cal E\}$. Let us denote by ${\bf f}^{(l)}\in\mathbb{R}^{n \times c}$ the feature tensor of the nodes $\mathcal{V}$ with $c$ channels at the $l$-th layer. We denote the adjacency matrix by $\bfA \in \mathbb{R}^{n \times n}$, where $\bfA_{ij} = 1$ if there exists an edge $(i,j) \in {\cal E}$ and 0 otherwise. We also define the diagonal degree matrix $\bfD$ where $\bfD_{ii}$ is the degree of the $i$-th node. 
We denote the adjacency and degree matrices with added self-loops by $\tilde \bfA$ and $\tilde \bfD$, respectively. 

In this paper we consider the node classification task, where the goal is to assign a class to each node in the graph. We denote the number of classes by $k$. Since not all nodes are labelled in our considered datasets, we denote the labelled set of vertices by $\mathcal{V}^{lab}\subset\mathcal{V}$, and the one-hot ground-truth labels by $\bfy \in \mathbb{R}^{\mid \mathcal{V}^{lab} \mid \times k}$. To present our approach, which also considers the unlabelled nodes, we denote the node classification prediction tensor by 
\begin{equation}\label{eq:y_hat}
\hat{\bfy} = {\rm SoftMax}(\bff^{out}) \in \mathbb{R}^{n \times k},
\end{equation}
where $\bff^{out}$ is the output last layer in the network.

\paragraph{GNN Backbones.}
In this paper we consider three popular backbones to demonstrate our ENC approach. Namely, we utilize GCN \citep{kipf2016semi}, GAT \citep{velickovic2018graph}, and GCNII \citep{chen20simple}.

\textbf{GCN}
defines a propagation operator
$\tilde{\bfP} = \tilde{\bfD}^{-\frac{1}{2}}\tilde{\bfA}\tilde{\bfD}^{-\frac{1}{2}}$, and its architecture is given by \begin{equation}
    \label{eq:general_gnn}
    \bff^{(l+1)} = \sigma(\tilde{\bfP} \bff^{(l)}\bfW^{(l)}),
\end{equation}
where $\bfW^{(l)}$ is a $1\times1$ convolution matrix, and $\sigma$ is a non-linear activation function. 

\textbf{GAT} defines the propagation operator according to the following edge weight:
\begin{equation}
    \label{eq:attentionCoefficients}
    \alpha_{ij}^{(l)} = \frac{\exp \big({\rm LeakyReLU} \big(\bfa^{(l)^{\top}} [\tilde{\bfW}^{(l)} \bff_i \mid \mid  \tilde{\bfW}^{(l)} \bff_j ]  \big) \big)}{\sum_{p \in \mathcal{N}_i} \exp \big({\rm LeakyReLU} \big(\bfa^{(l)^{\top}} [\tilde{\bfW}^{(l)} \bff_i \mid \mid  \tilde{\bfW}^{(l)} \bff_p ]  \big) \big)},
\end{equation}
where $\bfa^{(l)} \in \mathbb{R}^{2c}$ and $\tilde{\bfW}^{(l)} \in \mathbb{R}^{c \times c}$ are trainable parameters and $\mid \mid $ denotes channel-wise concatenation. 

By gathering $\alpha_{ij}^{(l)}$ for every edge $(i,j) \in \mathcal{E}$ into a propagation matrix $\bfS \in \mathbb{R}^{n \times n}$, a GAT layer reads:
\begin{equation}
    \bff^{(l+1)} = \sigma({\bfS^{(l)}} \bff^{(l)}\bfW^{(l)}).
\end{equation}

\textbf{GCNII} alters the propagation of GCN as follows:
\begin{equation}
\bfS^{(l)} (\bff^{(l)}, \bff^{(0)})=   (1-\alpha^{(l)}) \tilde{\bfP}\bff^{(l)} + \alpha^{(l)}\bff^{(0)},
\end{equation}
where $\alpha^{(l)} \in [0, 1]$ is hyper-parameter and $\bff^{(0)}$ are the features of the opening (embedding) layer. Then, a GCNII layer is given by:

\begin{equation}    
    \bff^{(l+1)} = \sigma(\beta^{(l)}\bfS^{(l)} (\bff^{(l)}, \bff^{(0)})\bfW^{(l)}  + (1-\beta^{(l)}) \bfS^{(l)} (\bff^{(l)},  \bff^{(0)})),
\end{equation}    
where $\beta^{(l)} \in [0,1]$ is a hyper-parameter.

\section{Method}
\label{sec:method}
The core idea of our method is that unlike the typical training scheme in GNNs, we treat every node in the data, regardless if it is labelled or not. Our experiments in Section \ref{sec:experiments} show that every node counts to improve performance, and thus we call our method ENC, which is summarized by the (three-term) extended objective function:
\begin{equation}
    \label{eq:totalObjective}
    \mathcal{L} = \mathcal{L}_{CE} + \alpha \mathcal{L}_{MI} + \beta \mathcal{L}_{TV} + \gamma \mathcal{L}_{CVG},
\end{equation}
where $\alpha \ , \beta\ , \gamma$ are non-negative hyper-parameters. Here, $\mathcal{L}_{CE}$ is the standard cross-entropy objective typically used in node classification tasks:
\begin{equation}
    \label{eq:crossEntropy}
    \mathcal{L}_{CE} = - \frac{1}{\mid \mathcal{V}^{lab}\mid}\sum_{i \in \mathcal{V}^{lab}}\sum_{s=1}^{k}  \bfy_{i,s} \log(\hat{\bfy}_{i,s}),
\end{equation} 
where $\hat{\bfy}$ is the prediction tensor defined in Eq. \eqref{eq:y_hat}. By definition, it considers only the labelled nodes $\mathcal{V}^{lab}$.

In the following, we present each of the proposed objectives, and later perform an hyper-parameter study in Section \ref{appendix:hyperparameters}. We further note that our method in Eq. \eqref{eq:totalObjective} does not require any architectural modifications to the baseline GNNs.

\subsection{Node-Class Mutual Information}
We define the mutual information objective as:
\begin{equation}
    \label{eq:mi_loss}
    \mathcal{L}_\mathrm{MI}=\frac{1}{\mid\mathcal{V}\mid}\sum_{i\in \mathcal{V}}\sum_{s=1}^{k}-\hat{\bfy}_{i,s}\log \hat{\bfy}_{i,s}+\lambda\sum_{s=1}^{k} \tilde{{\bfy}}_{s}\log\tilde{\bfy}_{s},
\end{equation}
where $\tilde{\bfy}_s=\frac{1}{\mid\mathcal{V}\mid} \sum_{i \in \mathcal{V}} \hat{\bfy}_{i,s}$ measures the mean prediction probability of the $s$-th class.
The first term in Eq. \eqref{eq:mi_loss} describes the entropy of the class prediction $[\hat{\bfy}_{i,s}]_{s=1}^{k}$ for each node $i\in\mathcal{V}$ individually, pushing the predicted classification vector to be deterministic. The second term considers the negative entropy of the mean class prediction probability, promoting the class-uniformity of the prediction. Note that $\mathcal{L}_{MI}$ considers all of the nodes $\mathcal{V}$, and in particular is not dependent on ground-truth labels as the standard cross-entropy loss. The value of $\lambda$ is a hyper-parameter and is dependent on the dataset, as a higher value will promote the network to equalize the number of predictions per class. In case $\lambda=1$, it was shown in \cite{bridle1991unsupervised} that Eq. \eqref{eq:mi_loss} maximizes the mutual information of the node and class features. Unless stated otherwise, in all experiments we set $\lambda=2$, and in Appendix \ref{appendix:hyperparameters} (in Tables \ref{table:semisupervisedHyperParams}-\ref{table:fullysupervisedHyperParameters}), we report the obtained accuracy using different values of $\lambda$.

\subsection{Total-Variation Regularization}
\label{sec:tvLoss}
The objective $\mathcal{L}_{TV}$ measures the node prediction discrepancy of neighbouring nodes, by utilizing the  Total-Variation (TV) \citep{rudin1992nonlinear} anisotropic smoothness prior that promotes correspondence between smooth regions while preserving boundaries in input features of the graph. To this end, we first define the gradient operator of the graph node features $\bff$ as:
\begin{equation}
    \label{eq:graphGradient}
    ({\nabla_{\mathcal{G}_{\bfw}}} {\bf f})_{ij} = (\bfw_i{\bf f}_i - \bfw_j{\bf f}_j) ,
\end{equation}
where nodes $i$ and $j$ are connected via the $(i,j)$-th edge, and $\bfw \in \mathbb{R}^{\mid\mathcal{V}\mid}$ is a node weight vector, which we discuss further below. Note that $\nabla_{{\cal G}_{\bfw}} \bff$ is a matrix of size $\mid \mathcal{E} \mid \times c$. With this definition, the standard TV regularization term is given by:
\begin{equation}
    \label{eq:tvSmoothingBasic}
    \frac{1}{\mid \mathcal{E \mid}}
    \sum_{(i,j)\in \mathcal{E}}\left\|(\nabla_{\mathcal{G}_{\bfw}} \hat{\bfy})_{ij}\right\|_1
    ,
\end{equation}
where $\hat{\bfy}$ is the prediction tensor defined in Eq. \eqref{eq:y_hat} and $\bff^{in} \in \mathbb{R}^{n \times c_{in}}$ are the input features tensor with $c_{in}$ channels. We augment Eq. \eqref{eq:tvSmoothingBasic} with an additional term that guides the TV regularization term to adhere to the boundaries in the input node features, as follows:
\begin{equation}
    \label{eq:tvSmoothing}
    \mathcal{L}_\mathrm{TV}=\frac{1}{\mid \mathcal{E \mid}}
    \sum_{(i,j)\in \mathcal{E}}\left(\left\|(\nabla_{\mathcal{G}_{\bfw}} \hat{\bfy})_{ij}\right\|_1 \exp^{-\|(\nabla_{\mathcal{G}_{\bfw}} \bff^{in})_{ij}\|_2^2/\sigma} 
    \right),
\end{equation}
where $\sigma$ is a scalar, set to 10 in our experiments. The $\exp$ term in Eq. \eqref{eq:tvSmoothing} was also proposed in the context of unsupervised image semantic segmentation tasks \citep{godard2017unsupervised}.

\paragraph{Total-Variation and the Dirichlet energy.}
In Eq. \eqref{eq:graphGradient} we set $\bfw_{i} = 1 / \sqrt{d_{i}+1}$, where $d_i$ is the degree of the $i$-th node, and ${\bf f}_i$ and ${\bf f}_j$ are the features of the $i$-th and $j$-th nodes, respectively.  
Note that the gradient operator is a mapping from the graph nodes to the edges, i.e., $\nabla_{\mathcal{G}_{\bfw}}: \mathcal{V} \longrightarrow \mathcal{E}$, and in particular its $\ell_2$ norm coincides with the Dirichlet energy:
\begin{equation}
    \label{eq:dirichletEnergy}
    E(\bff) = \sum_{(i,j)\in \mathcal{E}} \frac{1}{2} \textstyle{\left\|\frac{\bff_i}{\sqrt{(1+d_i)}} - \frac{\bff_j}{\sqrt{(1+d_j)}} \right\|_2^2} = \| \nabla_{\mathcal{G}_{\bfw}} \bff \|_2^2. 
\end{equation}
This observation uncovers an important nature of the proposed regularization technique --- it demands the similarity of the Dirichlet energy between input features and the node classification predictions. Nonetheless, using an $\ell_2$ norm to measure similarity is known not to respect the signal boundaries \citep{weickert1998anisotropic}, and typically smooth them. We therefore resort to the $\ell_1$ norm in Eq. \eqref{eq:tvSmoothing}, which was shown to be useful for color image processing when boundaries need to be preserved \citep{kong2014exclusive}.
The concept of preserving the Dirichlet energy is often used to avoid the over-smoothing phenomenon \citep{chen20simple,zhou2021dirichlet} by adding preserving terms to the \emph{network architecture}. Although the focus of this work is to obtain improved training, our experiments in Section \ref{sec:experiments} show that ENC also somewhat eases over-smoothing, albeit does not prevent it, as the baseline architectures are not changed.

\subsection{Cross-Validating Gradients}

The components in previous sections consider the inclusion of unlabelled nodes to the optimization objective. In this section we propose a third and final piece of our approach that seeks to improve the learning from the \emph{labelled} nodes. As discussed in Section \ref{sec:intro}, the typical training of GNNs  involves the minimization of the cross-entropy loss of the labelled nodes, as described in Eq. \eqref{eq:crossEntropy}. 
Here we use an additional mechanism to improve the training by 
requiring gradient consistency throughout the training process, which we obtain by demanding a cross-validation of the training procedure that we describe now. 

Let us denote a random disjoint  partition of the labelled nodes indices $p_1,\  p_2 \subset \mathcal{V}^{lab}$ where $p_1 \bigcup p_2 = \mathcal{V}^{lab}$ and $p_1 \bigcap p_2 = \emptyset$, that is uniformly drawn at each iteration during training. Let us consider the partial cross-entropy loss with respect to $p_1$ and $p_2$, i.e., ${L}_{CE}(\mathcal{V}^{lab}_{p_1})$ and $ {L}_{CE}(\mathcal{V}^{lab}_{p_2})$.
We wish that a gradient step computed with respect to the nodes $p_1$
will decrease the objective computed with respect to $p_2$ and vice versa. If this is not the case then the proposed direction may over-fit to a particular partition of points.
This discussion is at the core of the rational of using Generalized Cross Validation (GCV)
\citep{golub1979generalized,haber2000gcv,chung2010efficient} designed to prevent over-fitting for the mean square error loss and uses Jacobians with respect to the parameters. 

To avoid using Jacobians in the case of the cross-entropy loss, we propose to penalize the gradient steps that point to directions that fit one set of points $p_1$ and not the other $p_2$ by measuring their negative cosine similarity as follows
\begin{equation}
    \label{eq:crossValidatingGrads}
    \mathcal{L}_{CVG} = - \frac{{g_{\bftheta}^{p_1}} \cdot {g_{\bftheta}^{p_2}}}{\|{g_{\bftheta}^{p_1}}\| \|{g_{\bftheta}^{p_2}} \|},
\end{equation}
 where $g_{\bftheta}^{p_1} = \frac{\partial {L}_{CE}(\mathcal{V}^{lab}_{p_1})}{\partial \bftheta}$ and $g_{\bftheta}^{p_2} = \frac{\partial {L}_{CE}(\mathcal{V}^{lab}_{p_2})}{\partial \bftheta}$ are the  gradients of the partial cross-entropy losses with respect to the weights $\bftheta$, and $\cdot$ is the dot-product operation.
During training, we randomly generate equally-sized $p_1 \ , p_2$ using random permutations of the labelled nodes $\mathcal{V}^{lab}$.  
Each iteration can be thought of as a 2-fold cross validation and
it promotes steps that generalize the fit of the data. We show the positive effect of the stochasticity of $p_1 \ , p_2$ in Section \ref{sec:ablation}.

\subsection{Computational Costs}
\label{sec:computationalcost}
The losses $\mathcal{L}_{MI}$ and $\mathcal{L}_{TV}$ do not add significant computations, as they only consider the output of the network and perform simple operations. The loss $\mathcal{L}_{CVG}$ requires additional computations as it first computes the gradients $g_{\bftheta}^{p_1}$ and $g_{\bftheta}^{p_2}$. 
We measure and discuss the run-times and accuracy of the GNN baselines with our added objectives in Section \ref{sec:runtimes}. We can see that our approach offers a significant accuracy improvement over the baselines, at an insignificant additional cost using $\mathcal{L}_{MI}$ and $\mathcal{L}_{TV}$. Using all our losses from Eq. \eqref{eq:totalObjective}  (that is, also including $\mathcal{L}_{CVG}$ requires more computations, yielding further accuracy improvements.

\section{Experiments}
\label{sec:experiments}
\begin{figure*}
    \centering
    \includegraphics[width=0.32\textwidth]{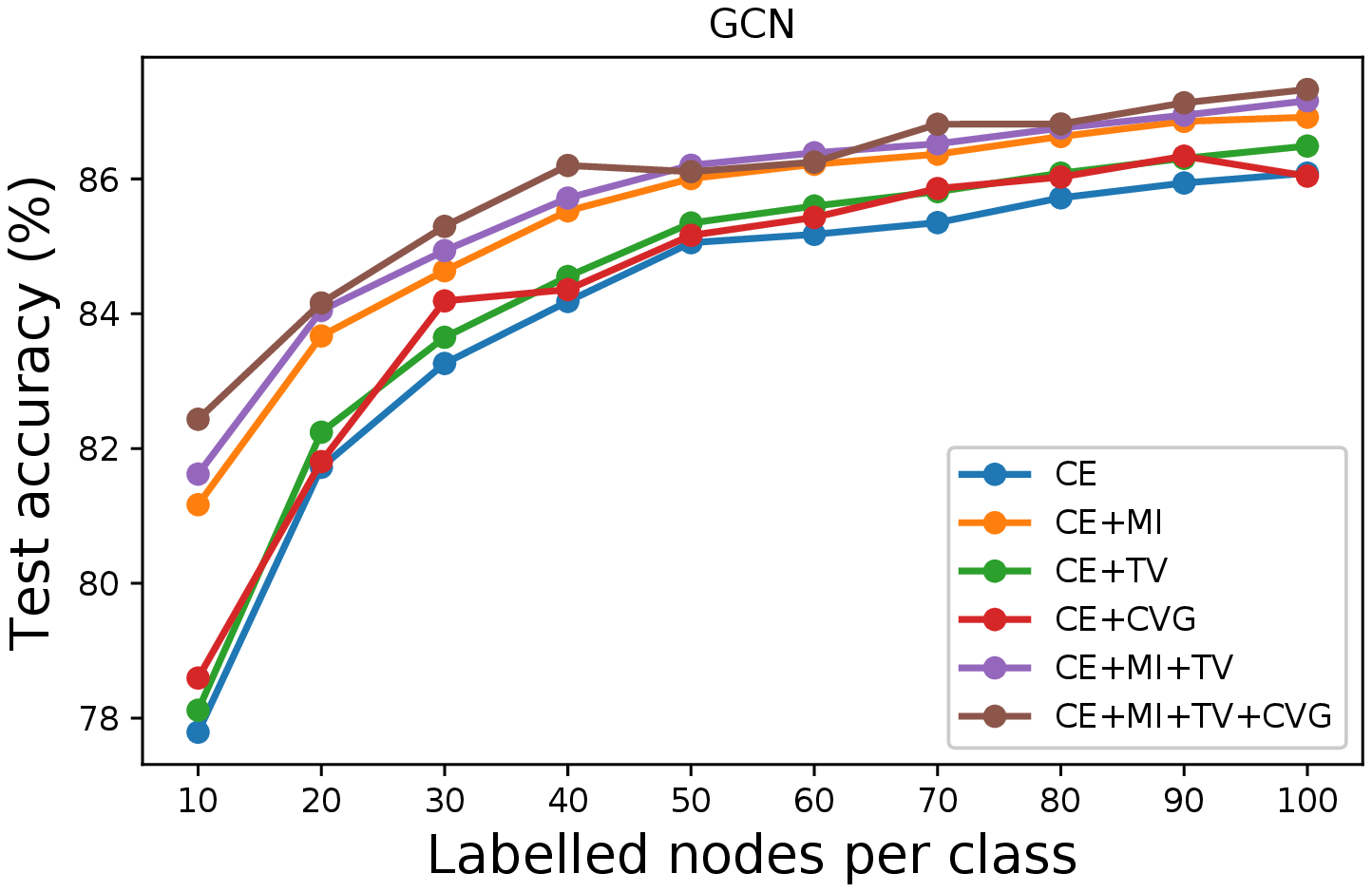}
    \includegraphics[width=0.32\textwidth]{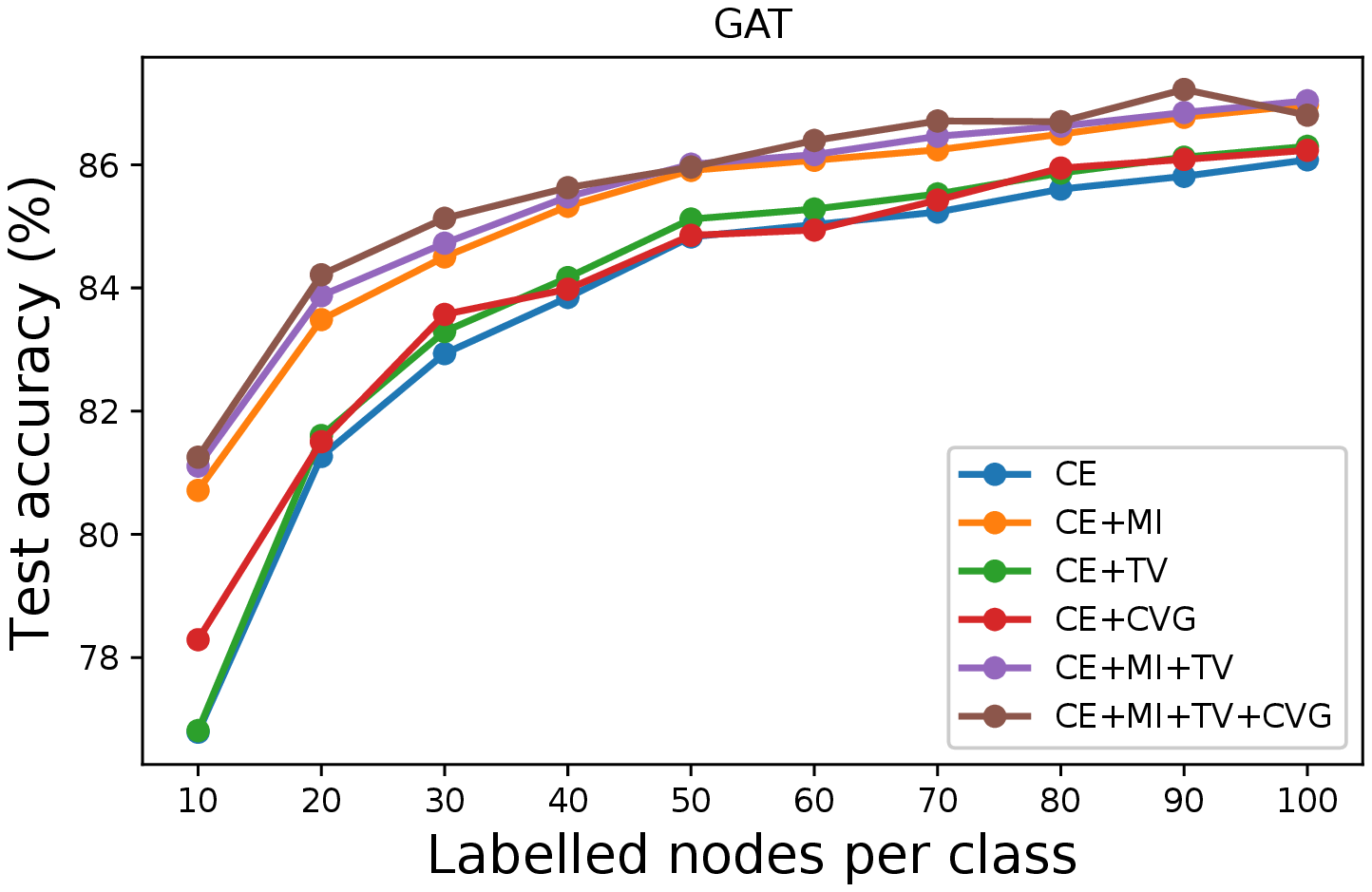}
    \includegraphics[width=0.32\textwidth]{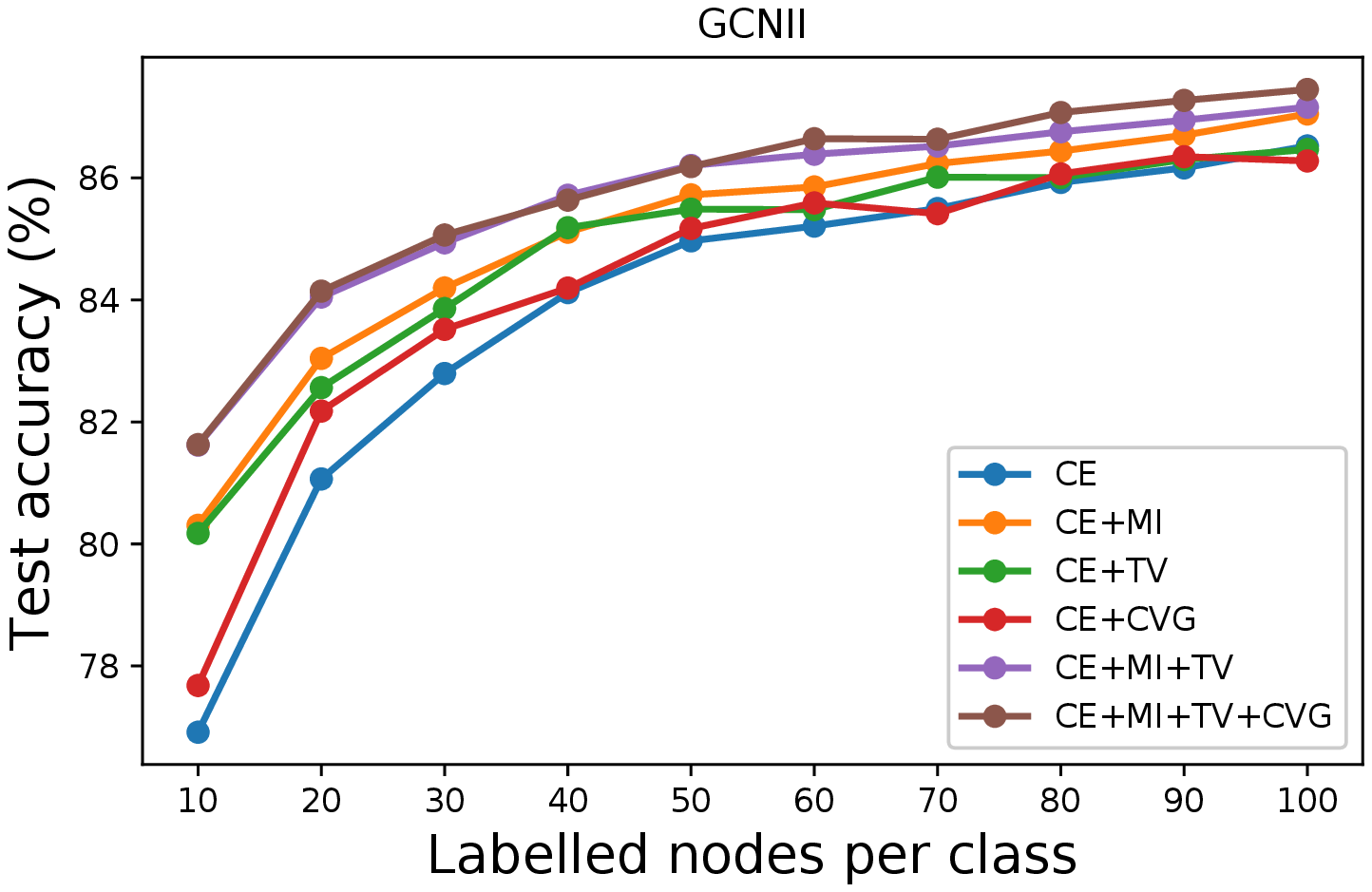}
    \caption{Cora test accuracy (\%) with a varying number of labels and 100 random splits per configuration. We compare the influence of the proposed losses. Our ENC approach (all losses together) consistently improves accuracy by an average of $2\%$. }
    \label{fig:objectivesInfluence}
\end{figure*}

We demonstrate our ENC on semi- and fully-supervised node classification followed by an ablation study.
In all experiments, we consider three popular baselines, namely GCN, GAT, and GCNII, to which we incorporate our ENC. We note that our method is general and can be incorporated to most GNNs, as it requires no architectural changes. A detailed description of the baselines that serve as backbones to our ENC approach is given in Section \ref{sec:notations}, and the general architecture is provided in Table \ref{table:nodeClassificationArch}, and  further information about the architecture is given in Section \ref{sec:architectures}. We use the Adam \citep{kingma2014adam} optimizer, and perform a grid search to choose the hyper-parameters. The range and selected values are reported in Section \ref{appendix:hyperparameters}. Our code is implemented using PyTorch \citep{pytorch} and PyTorch-Geometric \citep{pyg2019}, trained on an Nvidia Titan RTX GPU.

The statistics of the datasets used in our experiments are provided in Table \ref{table:datasets}. We show that for all the considered tasks and datasets, our ENC offers a consistent improvement over the baseline methods. For example, as shown in Table \ref{table:semisupervised_summary}, our ENC-GCN obtains 85.3\% accuracy on Cora, compared to 81.2\% using the standard GCN, an improvement of 4.1\%.

\begin{table}[t]
\begin{minipage}[t]{.48\linewidth}
    \caption{The GNN architecture in our experiments.}
  \label{table:nodeClassificationArch}
  \begin{center}
  \resizebox{1.0\textwidth}{!}{\begin{tabular}{lcc}
  \toprule
    Input size & Layer  &  Output size \\
    \midrule
    $n \times c_{in}$ & Dropout(p) & $n \times c_{in}$ \\
    $n \times c_{in}$ & $1\times1$ Convolution & $n \times c$ \\
    $n \times c$ & ReLU & $n \times c$ \\    
    $n \times c$ & $L \times $ ENC-GNN layers & $n \times c$ \\
    $n \times c$ & Dropout(p) & $n \times c$ \\
    $n \times c$ & $1\times1$ Convolution & $n \times k$ \\
    \bottomrule
  \end{tabular}}
\end{center}
  \end{minipage}
    \hspace{2pt}
  \begin{minipage}[t]{.48\linewidth}
  \caption{Datasets statistics.}
  \label{table:datasets}
  \resizebox{1.0\textwidth}{!}{
  \begin{tabular}{lcccc}
  \toprule
    Dataset & Classes & Nodes & Edges & Features  \\
    \midrule
    Cora & 7 & 2,708 & 5,429 & 1,433 \\
    Citeseer & 6 & 3,327  & 4,732 & 3,703 \\
    Pubmed & 3 & 19,717 & 44,338 & 500 \\
    Chameleon & 5 & 2,277 &  36,101 & 2,325\\
    Actor & 5 & 7,600 & 33,544 & 932  \\
    Squirrel & 5 & 5,201 & 198,493 &  2,089 \\
    Cornell & 5 & 183 & 295 & 1,703 \\
    Texas & 5 & 183 & 309 & 1,703 \\
    Wisconsin & 5 & 251 & 499 & 1,703 \\
    Ogbn-arxiv & 40 & 169,343  & 1,166,243 & 128 \\
    \bottomrule
  \end{tabular}}
  \end{minipage}
\end{table}

\subsection{Semi-Supervised Node Classification}
\label{sec:semisupervised}
We consider Cora \citep{mccallum2000automating}, Citeseer \citep{sen2008collective} and Pubmed  \citep{namata2012query}  datasets and their the standard, public training/validation/testing split as in \cite{yang2016revisiting}, with 20 nodes per class for training. 
We follow the training and evaluation scheme of \cite{chen20simple} and consider various GNN models like GCN, GAT, superGAT \citep{kim2020findSuperGAT}, APPNP \citep{klicpera2018combining}, JKNet \citep{jknet} , GCNII, GRAND \citep{chamberlain2021grand}, PDE-GCN \citep{eliasof2021pde} and EGNN\citep{zhou2021dirichlet} and superGAT \citep{kim2020findSuperGAT}. Additionally, we compare our approach with label propagation (LP) and regularization/augmentation based methods. For the former, we consider GCN-LPA \citep{wang2020unifying} and PTA \citep{dong2021equivalence}. For the latter, we compare with GAUG \citep{zhao2021data}, DropEdge \citep{Rong2020DropEdge:}, GraphVAT \citep{feng2019graph_adversarialtrain}, GRAND \citep{feng2020graph}, P-reg \citep{yang2021rethinking}, GraphMix \citep{verma2021graphmix}, NodeAug \citep{nodeaug_kdd20}, NASA \citep{bo2022regularizing_aaai22} and local augmentations (LA) by \cite{liu2022local}. We summarize the results in Table \ref{table:semisupervised_summary} where we see better or on par performance with other state-of-the-art methods, and a significant increase over the baselines of GCN, GAT and GCNII. For example, we obtain $83.8 \%$ accuracy on Pubmed using our ENC-GCNII compared to $80.3 \%$ with GCNII. We also provide a tSNE embedding of the obtained node predictions on Cora using GCN and our ENC-GCN in Figure \ref{fig:tsne}.

Additionally, we experiment with a varying number of layers, from 2 to 64, and report the results in Table \ref{table:semisupervised}. The evaluation of our method with a varying number of layers sheds light on the ability of our method to ease the over-smoothing phenomenon. While our method does not prevent the over-smoothing of the baseline method
(because it does not alter the architecture or feature aggregation scheme), we can see that compared to the over-smoothing baseline methods GCN and GAT, their counterparts ENC-GCN and ENC-GAT, respectively, show a slower accuracy degradation.

\begin{table}[t]
\center{
  \caption{Summary of semi-supervised node classification accuracy (\%)}
  \label{table:semisupervised_summary}
  \begin{tabular}{lccc}
  \toprule
    Method & Cora & Citeseer & Pubmed \\
    \midrule
    GCN & 81.1 & 70.8 & 79.0 \\
    GAT  & 83.1 & 70.8 & 78.5 \\
    GCNII & 85.5 & 73.4 & 80.3 \\
    \midrule
    ChebNet & 81.2 & 69.8 & 74.4 \\ 
    APPNP & 83.3 & 71.8 & 80.1  \\
    JKNET & 81.1 & 69.8 & 78.1 \\
    GRAND \citep{chamberlain2021grand} & 84.7 & 73.6 & 81.0 \\
    PDE-GCN & 84.3 & 75.6 & 80.6 \\
    EGNN & 85.7 & -- & 80.1 \\
    superGAT & 84.3 & 72.6 & 81.7 \\
    \midrule
    GCN-LPA & 82.8 & 72.3 & 78.6 \\
    PTA & 83.0 & 71.6 & 80.1 \\
    \midrule
    GAUG & 83.6 & 73.3 & 80.2 \\
    DropEdge & 82.8 & 72.3 & 79.6 \\
    GraphVAT & 82.9 & 73.8 & 79.5 \\
    GraphMix & 84.0 & 74.7 & 81.1 \\ 
    P-reg & 83.9  & 74.8 & 80.1 \\
    NodeAug & 84.3 & 74.2 & 81.5 \\
    GRAND \citep{feng2020graph} & 85.4 & 75.4 & 82.7 \\
    NASA & 85.1 & 75.5 & 80.2 \\
    
    LA-GCN  & 84.6 & 74.7 & 81.7 \\
    LA-GAT & 84.7 &  73.7 & 81.0 \\
    LA-GCNII & 85.7 & 74.1 & 80.6 \\
    LA-GRAND & 85.7  &  75.8 & 83.4 \\
 
    \midrule
    ENC-GCN  (Ours) & 85.3 & 75.5  & 81.9\\
    ENC-GAT  (Ours) & 85.2 & \textbf{75.8} & 82.6 \\
    ENC-GCNII  (Ours) & \textbf{86.0} & 75.0  & \textbf{83.8} \\

    \bottomrule
  \end{tabular}
}
\end{table}

\begin{table}[]
  \caption{Semi-supervised node classification accuracy ($ \%$). -- indicates not available results.}
  \label{table:semisupervised}
  \begin{center}
  \setlength{\tabcolsep}{1.5mm}{
  \begin{tabular}{llcccccc}
    \toprule
    \multirow{2}{*}{Dataset} & \multirow{2}{*}{Method} & \multicolumn{6}{c}{Layers} \\
                         &  & 2  & 4  & 8 & 16 & 32 & 64 \\
    \midrule
    Cora & GCN & \bf{81.1} & 80.4 & 69.5 & 64.9 & 60.3 & 28.7 \\
    & GAT & \textbf{83.1} & 81.3 & 74.5 & 68.2  & 58.9 & 34.1 \\
    & GCNII  & 82.2 & 82.6 & 84.2 & 84.6 & 85.4 & \bf{85.5} \\
    & GCN (Drop) & \bf{82.8} & 82.0 & 75.8 & 75.7 & 62.5 & 49.5 \\
    & JKNet (Drop) & -- & \bf{83.3} & 82.6 & 83.0 & 82.5 & 83.2 \\
    & GCNII*& 80.2 & 82.3 & 82.8 & 83.5 & 84.9 & \bf{85.3} \\
    & PDE-GCN\textsubscript{D} & 82.0 & 83.6 & 84.0 & 84.2 & 84.3 & \bf{84.3}  \\
    & EGNN & 83.2 & -- & -- & 85.4 & -- & \textbf{85.7}  \\
    & ENC-GCN & \textbf{85.3} & 84.4 & 83.3 & 80.9 & 75.4 & 60.6 \\
    & ENC-GAT & \textbf{85.2} & 84.5 & 82.5 & 80.4 & 75.1 & 66.3 \\
    & ENC-GCNII & 84.7 & 85.0 & 85.2 & 85.9  & 85.9 & \textbf{86.0}   \\
   \midrule
    Citeseer & GCN  & \bf{70.8} & 67.6 & 30.2 & 18.3 & 25.0 & 20.0 \\
    & GAT &  \textbf{70.8} & 68.6 & 55.7 & 31.2 & 22.0 & 20.9 \\
    & GCNII & 68.2 & 68.8 & 70.6 & 72.9 & \bf{73.4} & 73.4 \\
    & GCN (Drop) & \bf{72.3} & 70.6 & 61.4 & 57.2 & 41.6 & 34.4 \\ 
    & JKNet (Drop)  & -- & 72.6 & 71.8 & \bf{72.6} & 70.8 & 72.2 \\
    & GCNII* & 66.1 & 66.7 & 70.6 & 72.0 & \bf{73.2} & 73.1 \\
    & PDE-GCN\textsubscript{D} & 74.6 & 75.0 & 75.2 & 75.5 & \textbf{75.6} & 75.5 \\
    &ENC-GCN & \textbf{75.5} & 73.9 & 73.8 & 73.0 & 72.1 & 70.0   \\
    &ENC-GAT & \textbf{75.8} & 72.1 & 70.1 & 65.1 & 52.2 & 40.1 \\
    &ENC-GCNII & 72.5 & 73.4 & 73.8 & 74.0 & 74.6 & \textbf{75.0}\\

    \midrule
    Pubmed & GCN & \bf{79.0} & 76.5 & 61.2 & 40.9 & 22.4 & 35.3 \\
    & GAT & \textbf{78.5} & 76.4 & 69.8 & 50.1 & 41.0 & 42.3 \\
    & GCNII& 78.2 & 78.8 & 79.3 & \bf{80.2} & 79.8 & 79.7 \\
    & GCN (Drop) & \bf{79.6} & 79.4 & 78.1 & 78.5 & 77.0 & 61.5 \\
    & JKNet (Drop)  & -- & 78.7 & 78.7 & \bf{79.7} & 79.2 & 78.9 \\
    & GCNII* & 77.7 & 78.2 & 78.8 & \bf{80.3}& 79.8 & 80.1 \\
    & PDE-GCN\textsubscript{D} & 79.3 & \bf{80.6} & 80.1  & 80.4 & 80.2 & 80.3 \\
    & EGNN & 79.2 & -- & -- & 80.0 & -- & \textbf{80.1}  \\
    &ENC-GCN & \textbf{81.9} & 81.0 & 80.1 & 78.1 & 77.5 & 73.3 \\
    &ENC-GAT & \textbf{82.6} & 81.9 & 81.5 & 81.6 & 80.0 & 80.1 \\
    & ENC-GCNII & 81.0 & 81.4 & 81.8 & 82.6 & 83.2 & \textbf{83.8}  \\
    \bottomrule
  \end{tabular}}
\end{center}
\end{table}

\begin{table}[t]
\begin{minipage}[t]{.48\linewidth}
  \caption{Fully-supervised node classification accuracy ($ \%$) on \emph{homophilic} datasets}
  \label{table:homophilic_fully}   
  \center{
  \resizebox{1\linewidth}{!}{
  \begin{tabular}{lccc}
    \toprule
    Method & Cora & Cite. & Pub. \\
    Homophily & 0.81 & 0.80 & 0.74 \\
        \midrule
    GCN  & 85.77 & 73.68 & 88.13  \\
    GAT & 86.37 & 74.32 & 87.62 \\
    GCNII & 88.49  & 77.08 & 89.57  \\
    \midrule
    Geom-GCN-I & 85.19 & 77.99 & 90.05\\
    Geom-GCN-P & 84.93 & 75.14 & 88.09\\
    Geom-GCN-S & 85.27 & 74.71 & 84.75\\
    APPNP &  87.87 & 76.53 & 89.40 \\
    JKNet (Drop) & 87.46  & 75.96 & 89.45 \\
    WRGAT & 88.20 & 76.81 & 88.52 \\
    GCNII*  & 88.01 & 77.13 & \textbf{90.30}\\
    GGCN  & 87.95  & 77.14   & 89.15 \\
    H2GCN  & 87.87  & 77.11   & 89.49 \\
    GPRGNN & 87.95 & 77.13 & 87.54 \\
    \midrule 
    ENC-GCN (Ours) & 89.07 & 78.15 & 88.63 \\
    ENC-GAT (Ours) & 89.05 & 78.20 & 88.59 \\
    ENC-GCNII (Ours) & \textbf{90.01} & \textbf{79.34} &  89.35\\
    \bottomrule
  \end{tabular}}}
  \end{minipage} \hspace{5pt}
  \begin{minipage}[t]{.48\linewidth}
    \caption{Ogbn-arxiv node classification accuracy ($ \%$).}
  \label{table:ogbn_arxiv}   
  \center{
  \begin{tabular}{lc}
    \toprule
    Method & Acc. (\%) \\
        \midrule
    GCN  & 71.74  \\
    GAT & 71.59 \\
    GCNII & 72.74  \\
    \midrule
    APPNP & 71.82\\
    GATv2 & 71.87 \\
    EGNN &  72.70 \\
    APPNP &  72.23 \\
    GRAND & 72.70 \\
    \midrule
    ENC-GCN (Ours) & 72.95 \\
    ENC-GAT (Ours) & 73.38 \\
    ENC-GCNII (Ours) & \textbf{73.56}  \\
    \bottomrule
  \end{tabular}}
\end{minipage}

\end{table}

\begin{table}[t]
\centering
  \caption{Fully-supervised node classification accuracy ($ \%$) on \emph{heterophilic} datasets.}
  \label{table:heterophilic_fully}
   
  \begin{center}
  \begin{tabular}{lcccccc}
    \toprule
    Method & Squirrel & Actor &  Cham. & Corn. & Texas & Wisc. \\
    Homophily & 0.22 & 0.22 & 0.23 & 0.30  & 0.11 & 0.21 \\
        \midrule
    GCN  & 23.96 & 26.86 &  28.18 &  52.70 & 52.16 & 48.92 \\
    GAT & 30.03 & 28.45 & 42.93 & 54.32 & 58.38 & 49.41
    \\
    GCNII & 38.47 & 32.87 &   60.61  & 74.86 & 69.46 & 74.12 \\
    \midrule
    Geom-GCN-I & 38.32 & 29.09 &  60.31 & 56.76 & 57.58 & 58.24 \\
    Geom-GCN-P & 38.14 & 31.63 & 60.90 & 60.81  & 67.57 & 64.12 \\
    Geom-GCN-S & 36.24 & 30.30 &   59.96 & 55.68  & 59.73 & 56.67 \\
    JKNet (Drop) & 35.93 & 29.54 &  62.08  & 61.08 & 57.30 & 50.59  \\
    PairNorm & 50.44 & 27.40 & 62.74 & 58.92 & 60.27 & 48.43 \\
    GCNII*  & 39.92 & 33.61  & 62.48 & 76.49 & 77.84  & 81.57  \\
    GRAND & 40.05 & 35.62 &  54.67 & 82.16 & 75.68 & 79.41 \\ 
    WRGAT & 48.85 & 36.53 & 65.24 &  81.62 & 83.62 & 86.98 \\
    MagNet  &  --  & -- & --  & 84.30 & 83.30 & 85.70 \\ 
    GGCN  & \textbf{55.17} & \textbf{37.81} &  \textbf{71.14} & 85.68  & 84.86  &  86.86 \\
    H2GCN  & 36.48 & 35.70 & 60.11 & 82.70  & 84.86  &  87.65 \\
    GPRGNN & 31.61 & 34.63 & 46.58 & 80.27 & 78.38 & 82.94 \\
    FAGCN & 42.59 & 34.87 & 55.22 & 79.19 & 82.43 & 82.94 \\
    GraphCON-GCN & -- & -- & -- & 84.30 & 85.40 & 87.80 \\  
    GraphCON-GAT & --  & --  & -- & 83.20 & 82.20 & 85.70 \\  
    \midrule
    ENC-GCN (Ours) & 51.81 & 31.89 &  58.72& 73.14 & 61.08 & 59.80 \\
    ENC-GAT  (Ours) & 46.77 & 32.71 &   60.41 & 76.95  & 69.72 & 64.31 \\
    ENC-GCNII  (Ours)& 54.20 & 34.82 &  66.32 & \textbf{88.38} & \textbf{86.21} & \textbf{87.84} \\
    \bottomrule
  \end{tabular}
\end{center}
\end{table}

\subsection{Fully-Supervised Node Classification}
\label{sec:fullysupervised}
To further validate the efficacy of our method, we employ fully supervised node classification on 10 datasets. We examine our ENC-GCN, ENC-GAT and ENC-GCNII on Cora, Citeseer, Pubmed, Chameleon \citep{musae}, Squirrel, Actor, Cornell, Texas and Wisconsin using the 10 splits from \cite{Pei2020Geom-GCN:} with train/validation/test label split of $48 \%, 32\%, 20\%$ respectively, and report their average accuracy. In all experiments, 64 channels are used and a grid search is used to determine the hyper-parameters. To establish a strong baseline, we consider various methods, namely, GCN, GAT, Geom-GCN \citep{Pei2020Geom-GCN:}, APPNP, JKNet , WRGAT \citep{Suresh2021BreakingTL}, GCNII, PDE-GCN, DropEdge, H2GCN \citep{zhu2020beyondhomophily_h2gcn}, GGCN \citep{yan2021two}, MagNet \citep{zhang2021magnet}, GPRGNN \cite{chien2021adaptive}, FAGCN \cite{bo2021beyond}, and GraphCON \citep{rusch2022graph}. Additionally, we evaluate our ENC using the larger Ogbn-arxiv \citep{hu2020ogb} dataset using the official train/validation/test split in Table \ref{table:ogbn_arxiv}. To distinguish between homophilic and heterophilic datasets, we report the results of the former in Table \ref{table:homophilic_fully}, and of the latter in Table \ref{table:heterophilic_fully}. We see a significant improvement across all benchmarks and types of datasets compared to the baseline methods of GCN, GAT and GCNII. To measure the homophily score of the different datasets, we follow the definition in \cite{Pei2020Geom-GCN:}. 
Compared to recent methods like GraphCON, GGCN and H2GCN, our method reads better or similar accuracy while offering an appealing simplicity of keeping the baseline architectures, and changing only the training objective. For instance, our ENC-GCNII achieves $90.01\%$ accuracy on Cora compared to $87.95\%$ and $87.87 \%$ of GGCN and H2GCN, respectively. On a heterophilic dataset like Texas, our ENC-GCNII obtains an accuracy of $86.21\%$, while methods like GraphCON and H2GCN obtain $85.40 \%$ and $84.86\%$, respectively.

\subsection{Ablation Study}
\label{sec:ablation}
In this section we study the impact of the proposed objective functions.

\paragraph{Influence of the objectives}
As our ENC objective is comprised of several objective, it is important to delve on their contribution, individually and jointly, under different settings. We again use the GCN, GAT and GCNII architectures, and the Cora dataset. For a comprehensive study, we vary the number of labelled nodes per class from 10 to 100, with intervals of 10 and report the obtained test accuracy. To ensure statistically meaningful results and following observations from \cite{shchur2018pitfalls} regarding the evaluation of GNNs, for each experiment we report the average accuracy of 100 random splits with the respective number of labelled nodes. We present the results in Figure \ref{fig:objectivesInfluence}, where we can see that all of our objectives positively contribute to the obtained accuracy compared with the baseline case of using cross-entropy loss only. That is, we see that our ENC approach presented in Eq. \eqref{eq:totalObjective} obtains the best results across all considered settings. 

\begin{figure}[t]
    \centering
    \includegraphics[width=0.6\textwidth]{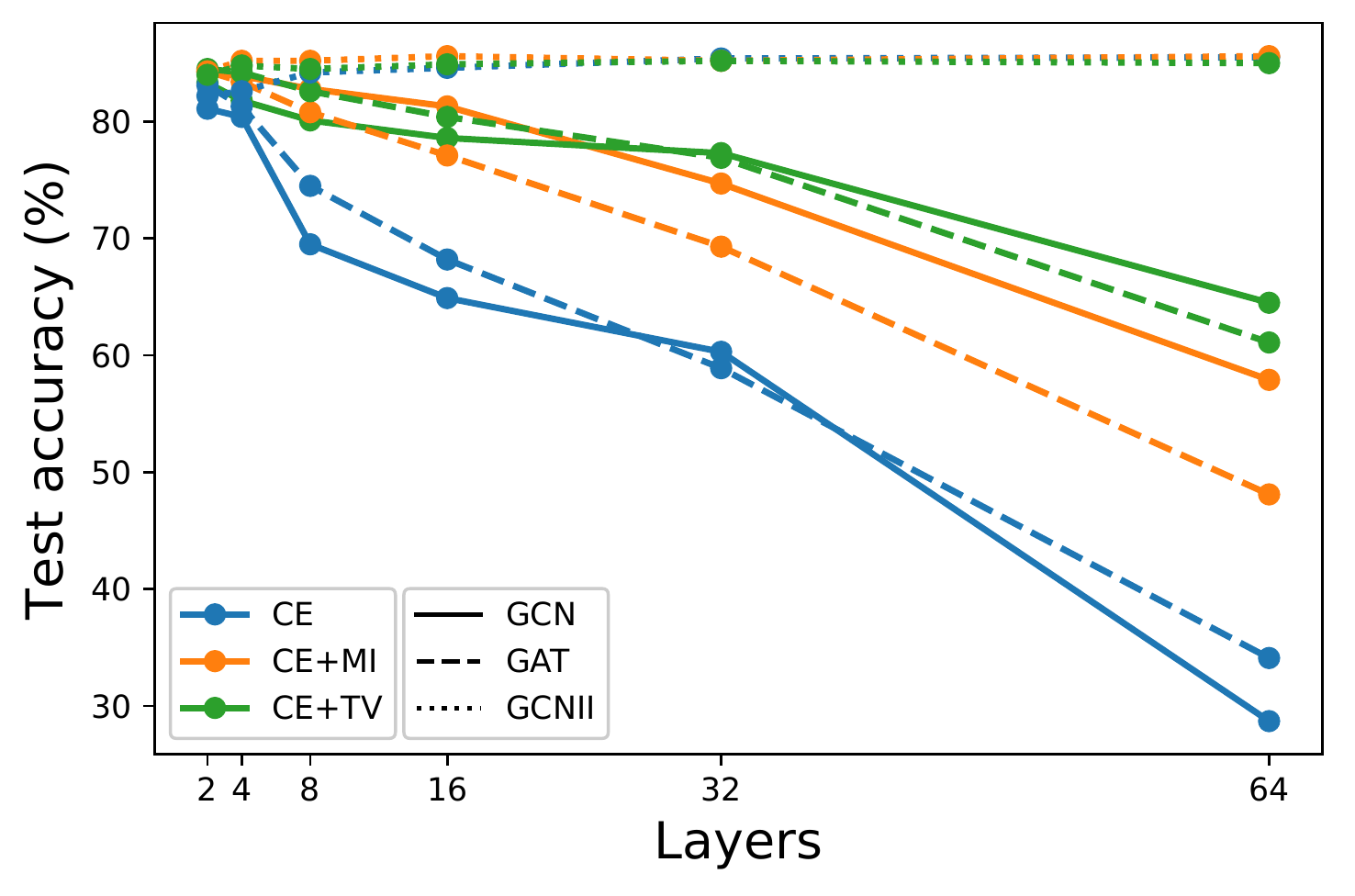}
    \caption{MI vs. TV loss on Cora using GCN, GAT, GCNII with a variable number of layers. TV loss obtains the best results with respect to the network's depth.}
    \label{fig:TVvsMI}
\end{figure}

In addition, although our ENC does not prevent over-smoothing, it is observed from Table \ref{table:semisupervised} that our ENC approach can somewhat ease the over-smoothing phenomenon. This is in spite of not changing the architecture, but only the loss function. To further investigate where this property stems from, we examined the performance on the public split of Cora with each of the proposed losses. We found that the $\mathcal{L}_{MI}$ and $\mathcal{L}_{TV}$ improve the performance of deep networks based on the GCN and GAT baselines, which are known to be over-smoothing \citep{measuringoversmoothing,Zhao2020PairNorm:}. Specifically, we find that the TV loss yields the best results as an individual loss with respect to the depth of the networks. We found that using the CVG loss does not achieve a similar effect. For a complete comparison, we also report the results with GCNII as baseline. The results are reported in Figure \ref{fig:TVvsMI}. 

\begin{figure}[t]
    \centering
    \includegraphics[width=0.6\textwidth]{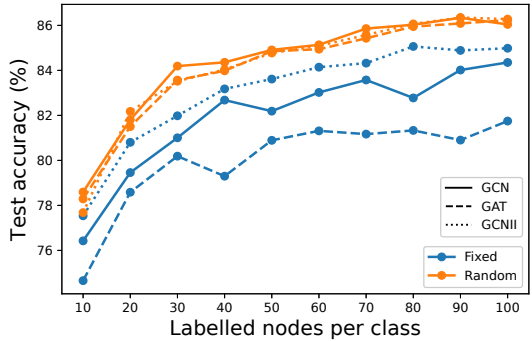}
    \caption{The effect of randomly sampling $p_1$ and $p_2$ demonstrated on the Cora dataset using GCN, GAT and GCNII.}
    \label{fig:fixedVsRandom}
\end{figure}

\paragraph{Fixed vs. Random nodes partition}
We study the effect of randomly partitioning the set of labelled nodes $\mathcal{V}^{lab}$ to $p_1 \ , p_2$ at every iteration compared to fixing $p_1 \, p_2$ throughout the training stage on the Cora dataset, using GCN, GAT and GCNII. In the case of the latter, we report the average accuracy obtained by 10 random initializations of $p_1$ and $p_2$ to ensure the significance of the results. We present the results in Figure \ref{fig:fixedVsRandom}. We can immediately see a large performance gap between the two choices, leading us to employ the random sampling of the node partition at every iteration in our experiments.

\paragraph{TV vs. P-reg terms}
The P-reg \citep{yang2021rethinking} regularization leverages on the known Laplacian regularization \citep{smola2003kernels}, to improve the training of GNNs by promoting smooth node predictions. While it was demonstrated to be effective on homophilic datasets like Cora, Citeseer and Pubmed, it is interesting to find whether such a strategy can be beneficial in heterophilic dataets like Cornell, Texas and Wisconsin. Intuitively, one may expect such a method to perform worse, as it demands the similarity of predicted labels and their smoothing by the normalized adjacency matrix, as follows:
\begin{equation}
    \label{eq:laplacianReg}
    \mathcal{L}_{P-reg} = \| \hat{\bfy} - \tilde{\bfD}^{-1}\tilde{\bfA} \hat{\bfy}   \|_2^2.
\end{equation}
In contrast, our TV regularization in Eq. \eqref{eq:tvSmoothing} suggests to promote learning predictions that are both smooth and adherent to the natural boundaries of the input features. We therefore expect that for heterophilic datasets, our method will perform better. To this end we experiment with GCN joint with each of the regularizations, one at a time, and report the test accuracy obtained on Cora, Citeseer and Pubmed, Cornell, Texas and Wisconsin using the splits from \cite{Pei2020Geom-GCN:} in Table \ref{table:TVvsLaplacianReg}. We find that for homophilic datasets, both methods improve the baseline GCN. However, for heterophilic datasets, employing P-reg regularization can harm the accuracy, further highlighting the contribution of the TV regularization as a smoothing but also boundary preserving regularizer.

\begin{table}[ht!]
    \centering
    \caption{Total variation vs. P-reg regularization applied to GCN. Metric is accuracy ($\%$). Hom. denotes Homophily. TV regularization improves accuracy both on homophilic and heterophilic datasets.}
  \label{table:TVvsLaplacianReg}
  \begin{tabular}{lccccc}
    \toprule
    \multirow{2}{*}{Dataset} & \multirow{2}{*}{Hom.} & \multirow{2}{*}{GCN} & GCN+ & GCN+ & GCN+ \\& & & TV (Eq. \eqref{eq:tvSmoothingBasic}) & P-reg. & TV (Eq. \eqref{eq:tvSmoothing}, ours) \\
    \hline
    Cora & 0.81 & 85.77 & 87.47 &  87.42 & \textbf{88.17} \\
    Citeseer& 0.80 & 73.68 & 77.02 &  76.96 & \textbf{77.10} \\
    Pubmed & 0.74  & 88.13 & 87.14 &  \textbf{88.34} & \textbf{88.34}  \\
    \hline
    Cornell & 0.30 & 52.70 & 63.22 &  61.25 & \textbf{66.71}  \\
    Texas &  0.11 & 52.16 & 
 56.80& 50.21 & \textbf{58.16} \\
    Wisconsin &  0.21 &48.92 & 57.83&  47.88 & \textbf{58.39}  \\
    \bottomrule
  \end{tabular}
    
    \end{table}

\section{Conclusion}
In this paper we propose an orthogonal path to the recent advances in GNNs. While most methods focus on the improvement of GNN architectures and relying on the standard cross-entropy loss, we show that by including our ENC objectives in the optimization process leads to major improvements of baseline methods like GCN, GAT and GCNII. Our method often achieves better or similar results to other state-of-the-art methods that are more complex. We motivate our objectives by adapting knowledge and concepts from fields like Computer Vision, Image Processing and Optimization methods that are often found in CNNs but not in GNNs, and validate their efficacy in our extensive set of experiments. Our method is quite general, and we deem that it will also be beneficial for training future GNN architectures.

\bibliography{sn-bibliography}

\section{Appendix}
\label{sec:implementation_details}
We now elaborate on the specific hyper-parameters and their influence, followed by a run-time discussion and measurement.

\subsection{Architecture}
\label{sec:architectures}
All our network architectures consist of an opening (embedding) layer ($1
\times 1$ convolution), a sequence of ENC-GNN (see details below for specific aggregation rules for GCN/GAT/GCNII) layers, and a closing (classifier) layer ($1
\times 1$ convolution).
We have a single type of architecture, based on the grand scheme of GCN for node classification tasks. The only difference between our ENC-GCN/ENC-GAT/ENC-GCNII is the backbone of the GNN. We report the architecture in Table \ref{table:nodeClassificationArch}. In what follows, we denote by $c_{in}$ and $k$ the input and output channels, respectively, and $c$ denotes the number of features in hidden layers. We initialize the embedding and classifier layers with the Glorot \citep{glorot2010understanding}
initialization, and $\bfW^{(l)}$ from Eq. \eqref{eq:general_gnn} is initialized with an identity matrix of shape $c \times c$. We denote the number of GNN layers by $L$, and the dropout probability by $p$.
It is important to note that our ENC-GNNs do not vary the architectures of considered baselines, but only change the objective function.

\subsection{Hyper-parameters}
\label{appendix:hyperparameters}
We now provide the selected hyper-parameters in our experiments, chosen by a grid-search. We denote the learning rate of our GNN layers by $LR_{GNN}$, and the learning rate of the $1\times 1$ opening and closing layers by $LR_{oc}$.
Also, the weight decay of GNN layers is denoted by $WD_{GNN}$, and the weight decay of the opening and closing layers is denoted by $WD_{oc}$. $c$ denotes the number of hidden channels.
Our search space in all experiments for the hyper parameter is as follows: $LR_{GNN} , LR_{oc} \in [1e-5, 0.1]$, $WD_{GNN},WD_{oc} \in [0, 0.1]$, and $\alpha, \beta, \gamma \in [0.01,5]$.
Our hyperparameters for the semi- and fully-supervised node classification experiments are given in Table \ref{table:semisupervisedHyperParams} and \ref{table:fullysupervisedHyperParameters}, respectively. In our ablation studies we followed the same hyper-parameters from Table \ref{table:semisupervisedHyperParams}-\ref{table:fullysupervisedHyperParameters} according to the respective splits used in each experiment.

\begin{figure}[t]
    \centering
    \includegraphics[width=0.6\textwidth]{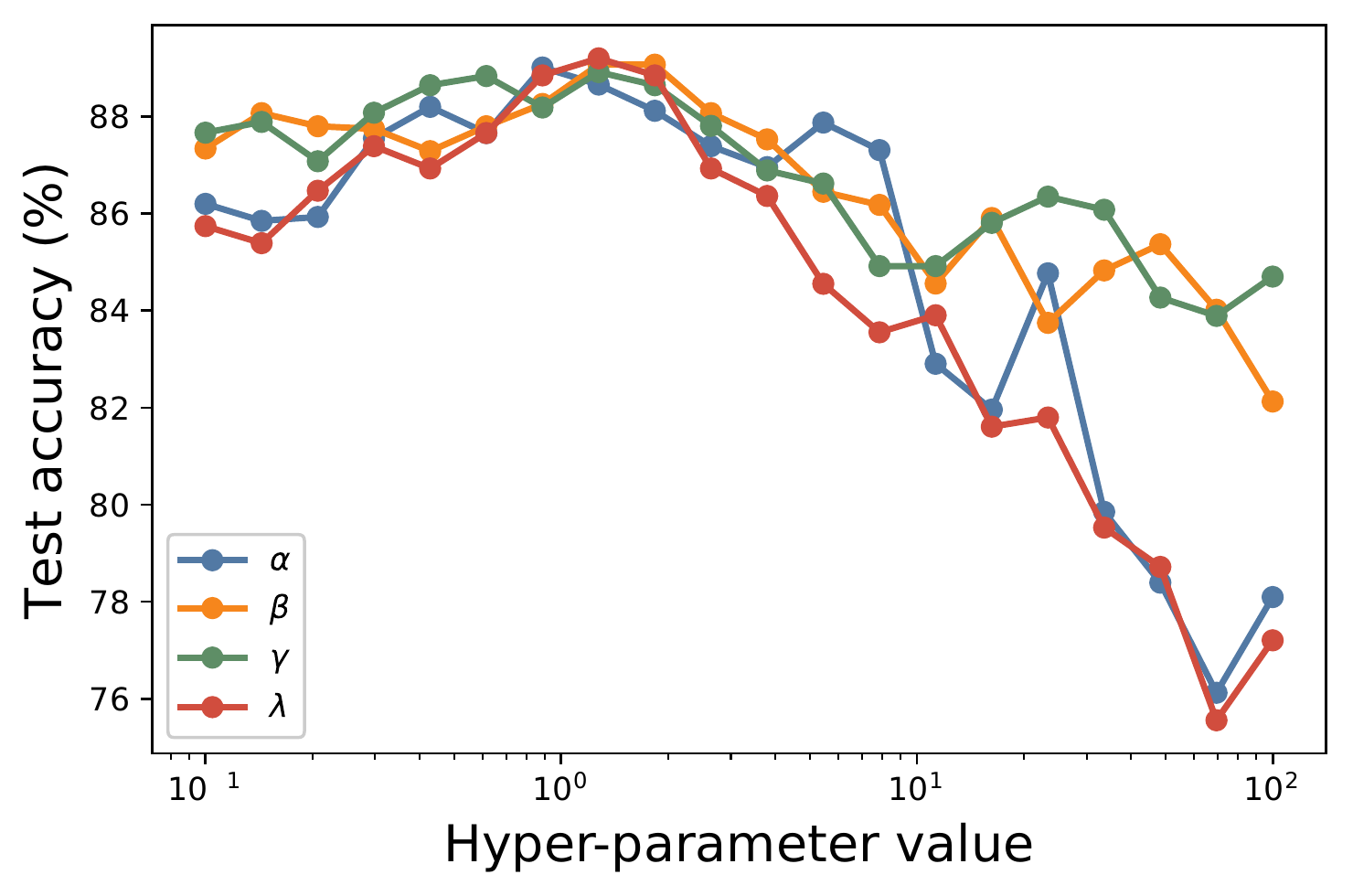}
    \caption{Hyper-parameter study on Cora.  Each hyper-parameter is varied on a log-space while the rest are fixed to 1 for $\alpha, \beta, \gamma$ and 2 for $\lambda$.}
    \label{fig:hyperparamStudy}
\end{figure}

Also, in Figure \ref{fig:hyperparamStudy} we provide a hyper-parameters study of $\alpha, \beta, \gamma$ and $\lambda$ from Eq. \eqref{eq:totalObjective} and Eq. \eqref{eq:mi_loss}, respectively, using the Cora dataset. We report the average test accuracy using the 10 splits from \cite{Pei2020Geom-GCN:}. For each loss we only change its hyper-parameter, and keep the rest equal to 1, besides $\lambda$ which when not considered, is fixed to 2. According to the results of this study, we decided to search our hyper-parameters $\alpha,\beta,\gamma$ in the range of [0.1,5] and fix $\lambda=2$.

\begin{table}[]
  \caption{Training and inference GPU run-times [ms] on Cora.}
  \small
  \label{tab:runtimes}
  \begin{center}
  \begin{tabular}{lccc}
  \toprule
   Method  & Training & Inference & Accuracy \\
    \midrule
    GCN & 12.58 & 5.88 & 81.1 \\
    GAT  & 19.51  & 7.14 & 83.1\\
    GCNII & 12.86 & 5.99 & 80.1 \\
    \midrule
    GCN + $\mathcal{L}_{MI}$ + $\mathcal{L}_{TV}$& 13.01 & 5.88 & 84.9\\
    GAT + $\mathcal{L}_{MI}$ + $\mathcal{L}_{TV}$ & 19.88 & 7.14& 84.7  \\
    GCNII  + $\mathcal{L}_{MI}$ + $\mathcal{L}_{TV}$ & 13.32 & 5.99 & 84.3 \\
    \midrule
    ENC-GCN & 23.12 &5.88 & 85.3\\
    ENC-GAT  & 41.61 & 7.14 & 85.2\\
    ENC-GCNII & 24.27 & 5.99 & 84.7 \\
    \bottomrule
  \end{tabular}
  \end{center}
\end{table}

\begin{table*}[t]
  \caption{Semi-supervised node classification hyper-parameters.}
  \small{
\label{table:semisupervisedHyperParams}
  \begin{center}
  \resizebox{1.0\columnwidth}{!}{
  \begin{tabular}{llcccccccccc}
  \toprule
    Method & Dataset & $LR_{GNN}$ & $LR_{oc}$ & $WD_{GNN}$ & $WD_{oc}$ &  $c$ & $p$ & $\alpha$ & $\beta$ & $\gamma$   \\
    \midrule
    ENC- & Cora & 1e-3 & 0.01 &  1e-5 & 1e-5 &  64 & 0.6 & 1.0 & 2.0 & 1.0 \\
    GCN &Citeseer & 1e-4 & 5e-3  &  5e-3 & 1e-4 & 256 & 0.7  & 0.8 & 2.0 & 1.0  \\
    & Pubmed &  0.01 & 5e-4 & 1e-4 & 1e-4 & 256 & 0.5 & 0.6 & 2.0 & 1.0  \\
    \midrule   
    ENC-  & Cora & 1e-3 & 0.01 & 1e-4 & 1e-4 &  64 & 0.6 & 0.8 & 1.0 & 1.0 \\
    GAT &Citeseer & 0.01 & 0.01 & 1e-3 & 1e-4 & 256 & 0.7 & 1.0 & 2.0 & 2.0   \\
    & Pubmed & 1e-3 & 0.01 & 1e-3  & 1e-4 & 256 & 0.5 & 1.0 & 1.0 & 1.0 \\
    \midrule   
    ENC-  & Cora & 5e-3 & 0.01 & 1e-4 & 1e-5 &  64 & 0.6 & 0.8 & 1.6 & 1.2\\
    GCNII&Citeseer & 5e-3 & 5e-3 & 1e-5 & 1e-4 & 256 & 0.7 & 1.2 & 1.0 & 0.8   \\
    & Pubmed & 5e-3 & 1e-3 & 0.05  & 1e-4 & 256 & 0.5 & 0.6 & 1.6 & 1.0\\
    \bottomrule
  \end{tabular}}
\end{center}}
\end{table*}

\begin{table*}[t]
  \caption{Fully-supervised node classification hyper-parameters.}
  \small
  \label{table:fullysupervisedHyperParameters}
  \begin{center}
  \resizebox{1.0\columnwidth}{!}{\begin{tabular}{llcccccccccc}
  \toprule
   Method &  Dataset & $LR_{GNN}$ & $LR_{oc}$ & $WD_{GNN}$ & $WD_{oc}$ & $c$ & $p$ & $\alpha$ & $\beta$ & $\gamma$  \\
    \midrule
    ENC- & Cora & 1e-3 & 0.01 &  5e-3 & 53-4 & 64 & 0.5 &  0.6 & 1.0 & 1.0    \\
    GCN &Citeseer & 1e-3 & 0.01 & 5e-3  & 1e-4 & 64 & 0.5 & 0.6 & 0.6 & 1.0  \\
    &Pubmed & 0.01 & 0.01 & 0.01 & 5e-4 & 64 & 0.5 & 1.2 & 0.6 & 1.0  \\
    &Chameleon & 1e-4 & 0.05 & 5e-5 & 0 & 64 & 0.5 & 2.0 & 2.0 & 1.2\\
    &Actor (Film) & 1e-3 & 0.05 & 0.05 & 1e-4 & 64 & 0.5 & 1.2 & 4.4 & 1.8 \\
    & Squirrel & 5e-3 & 0.05 & 1e-5 & 0 & 64 & 0.5 & 4.2 & 2.2 & 2.6 \\
    &Cornell & 0.05 & 1e-4 & 1e-4 & 1e-4 & 64 & 0.5 & 1.0 & 1.0 & 2.0 \\
    &Texas & 1e-3 & 0.05 & 1e-5 &  1e-5 & 64 & 0.5  & 0.6 & 2.0 & 2.0\\
    &Wisconsin & 1e-3 & 0.05 & 5e-3 & 5e-4 & 64  & 0.5 & 1.0  & 1.0 & 1.0  \\
    &Ogbn-arxiv & 0.01 & 0.01 & 0 & 0 & 256 & 0 & 2.0 & 1.0 & 0.8  \\
    \midrule
    ENC- & Cora & 0.01 & 0.01 &  1e-3 & 1e-4 & 64 & 0.5 & 1.6 & 1.2 & 1.0   \\
    GAT&Citeseer & 1e-3 & 0.05 & 1e-5  & 5e-4 & 64 & 0.5 & 1.8 & 1.2 & 1.0  \\
    &Pubmed & 1e-3 & 1e-3 & 1e-4 & 1e-5 & 64 & 0.5 & 1.6 & 0.6 & 0.6 \\
    &Chameleon & 1e-3 & 0.01 & 5e-4 & 1e-5 & 64 & 0.5 & 1.0 & 0.8 & 1.0 \\
    &Actor (Film) & 0.05 & 1e-4 & 1e-5 & 5e-4 & 64 & 0.5 & 2.0 & 1.0 & 1.0 \\
    & Squirrel & 0.05 & 1e-3 & 0 & 1e-5 & 64 & 0.5 & 1.0 & 1.2 & 0.8\\
    &Cornell & 0.01 & 0.05 & 0.01 & 0 & 64 & 0.5  & 1.8 & 1.0 & 1.4\\
    &Texas & 1e-3 & 0.05 & 5e-4 &  1e-4 & 64 & 0.5 & 1.2 & 0.6 & 0.8 \\
    &Wisconsin & 1e-3 & 0.05 & 1e-3 & 1e-4 & 64  & 0.5 & 2.6 & 0.8 & 1.4  \\
    &Ogbn-arxiv & 0.01 & 0.01 & 0 & 0 & 256 & 0 & 1.4 & 1.8 & 1.0 \\
    \midrule
    ENC- & Cora & 0.01 & 0.01 &  0.05 & 1e-4 & 64 & 0.5 & 0.8 & 0.8 & 1.0   \\
    GCNII &Citeseer & 1e-4 & 0.01 & 1e-3 & 5e-4 & 64 & 0.5 & 2.0 & 1.0 & 1.2  \\
    &Pubmed & 0.05 & 0.05 & 0.05 & 0 & 64 & 0.5 & 3.0 & 1.0 &  1.4 \\
    &Chameleon & 0.01 & 0.01 & 1e-4 & 1e-5 & 64 & 0.5 & 0.6 & 0.8 & 1.0\\
    &Actor (Film) & 0.05 & 0.01 & 0.01 & 1e-4 & 64 & 0.5 & 1.6 & 4.0 & 1.4 \\
    & Squirrel & 0.01 & 0.01 & 1e-5 & 1e-5 & 64 & 0.5 & 0.8 & 1.0 & 1.0 \\
    &Cornell & 0.01 & 0.05 & 0.01 & 0 & 64 & 0.5  & 1.0 & 1.2 & 0.8\\
    &Texas & 0.01 & 0.05 & 1e-3 &  1e-3 & 64 & 0.5 & 1.6 & 0.8 & 1.2 \\
    &Wisconsin & 0.01 & 0.01 & 5e-4 & 5e-3 & 64  & 0.5 & 1.0 & 4.0 & 1.6 \\
    &Ogbn-arxiv & 0.01 & 0.01 & 0 & 0 & 256 & 0 & 1.0 & 2.0 & 1.0 \\
    \bottomrule
  \end{tabular}}
\end{center}
\end{table*}

\subsection{Run-times}
\label{sec:runtimes}
Following the computational cost discussion from Section \ref{sec:computationalcost} in the main paper, we present in Table \ref{tab:runtimes} the measured training and inference times of the baseline GCN, GAT and GCNII with 2 layers. All the measurements were done on a single Nvidia Titan RTX GPU with 24GB of memory. We see that the incorporation of $\mathcal{L}_{MI}$ and $\mathcal{L}_{TV}$ of our ENC to the baseline methods requires an insignificant addition of time, at the return of a significantly better test accuracy on the public semi-supervised split of Cora. We note that further accuracy gain can be achieved by incorporating the full ENC method (i.e., also including $\mathcal{L}_{CVG}$) which requires more computations, as discussed in Section \ref{sec:computationalcost}. Also, it is important to note that the sole difference is in the \emph{training} time, while \emph{inference} times are identical. This is because our ENC-GNN method does not change the architecture of the network but only its objective, yielding the exact same inference times as the baseline models.

\end{document}